\documentclass[10pt,twocolumn,letterpaper]{article}

\usepackage{iccv}
\usepackage{times}
\usepackage{epsfig}
\usepackage{graphicx}
\usepackage{amsmath}
\usepackage{enumitem}
\usepackage{comment}
\usepackage{bm}
\usepackage{amssymb}
\usepackage{bbm}
\usepackage{math}
\usepackage{cases}
\usepackage{xcolor}
\definecolor{Orange}{rgb}{0.9,0.5,0}
\definecolor{NavyBlue}{rgb}{0.1, 0.4, 0.8}
\definecolor{Magenta}{rgb}{0.8, 0.1, 0.6}
\definecolor{mypink2}{RGB}{219, 48, 122}

\usepackage{floatrow}
\PassOptionsToPackage{hyphens}{url}

\newfloatcommand{capbtabbox}{table}[][\FBwidth]

\usepackage[pagebackref=true,breaklinks=true,letterpaper=true,colorlinks,bookmarks=false]{hyperref}

\usepackage{booktabs}
\usepackage{tabularx}
\usepackage{multirow}
\newcommand{\specialcell}[2][l]{%
\begin{tabular}[#1]{@{}l@{}}#2\end{tabular}}

\usepackage{cite}
\usepackage{cleveref}
\iccvfinalcopy 

\ificcvfinal\fi

\begin{document}

\title{Click to Move: Controlling Video Generation with Sparse Motion}

\author{
  Pierfrancesco Ardino\textsuperscript{1,2},
  Marco De Nadai\textsuperscript{2},
  Bruno Lepri\textsuperscript{2},
  Elisa Ricci\textsuperscript{1,2},
  St{\'e}phane Lathuili{\`e}re\textsuperscript{3}\vspace{0.2cm}
\\
  \textsuperscript{1}University of Trento \quad
  \textsuperscript{2}Fondazione Bruno Kessler \\
  \textsuperscript{3}LTCI, T{\'e}l{\'e}com Paris, Institut Polytechnique de Paris
\vspace{-1em}
}

\maketitle
\ificcvfinal\thispagestyle{empty}\fi

\begin{abstract}
This paper introduces Click to Move (C2M), a novel framework for video generation where the user can control the motion of the synthesized video through mouse clicks specifying simple object trajectories of the key objects in the scene. Our model receives as input an initial frame, its corresponding segmentation map and the sparse motion vectors encoding the input provided by the user. It outputs a plausible video sequence starting from the given frame and with a motion that is consistent with user input. Notably, our proposed deep architecture incorporates a Graph Convolution Network (GCN) modelling the movements of all the objects in the scene in a holistic manner and effectively combining the sparse user motion information and image features. 
Experimental results show that C2M outperforms existing methods on two publicly available datasets, thus demonstrating the effectiveness of our GCN framework at modelling object interactions. 
The source code is publicly available at \url{https://github.com/PierfrancescoArdino/C2M}.

\end{abstract}

\section{Introduction}


Recent years have witnessed several breakthroughs in the generation of high dimensional data such as images~\cite{choi2019stargan, dong2020fashion, Liu_2021_CVPR} or videos~\cite{tulyakov2018mocogan,vondrick2017generating}. However, most practical and commercial applications require to control generated visual data on inputs provided by the user.
For instance, in image manipulation, photo editing software~\cite{Adobe} applies deep learning models to allow users to change portions of an image~\cite{yu2019region, shen2020interpreting, nazeri2019edgeconnect}. 

Regarding videos, several possible ways to control the generated sequences have been considered. 
For instance, the generation of frames can be conditioned on simple categorical attributes \cite{he2018probabilistic}, short sentences \cite{li2018video} or sound \cite{tsuchiya2019generating}.
An interesting recent research direction comprises works that attempt to condition the video generation process providing motion information as input \cite{tulyakov2018mocogan,wiles2018x2face,siarohin2019animating,siarohin2020first}. These approaches allow to generate videos of moving faces \cite{wiles2018x2face}, human silhouettes and, in general, of arbitrary objects \cite{tulyakov2018mocogan,siarohin2019animating,siarohin2020first}. 
However, these works mainly deal with videos depicting a single object. It is indeed extremely more challenging to animate images and generate videos when multiple objects are present in the scene, as there is no simple way to disentangle the information associated with each object and easily model and control its movement. 

\begin{figure}[t]	
	\centering
	\includegraphics[width=\linewidth]{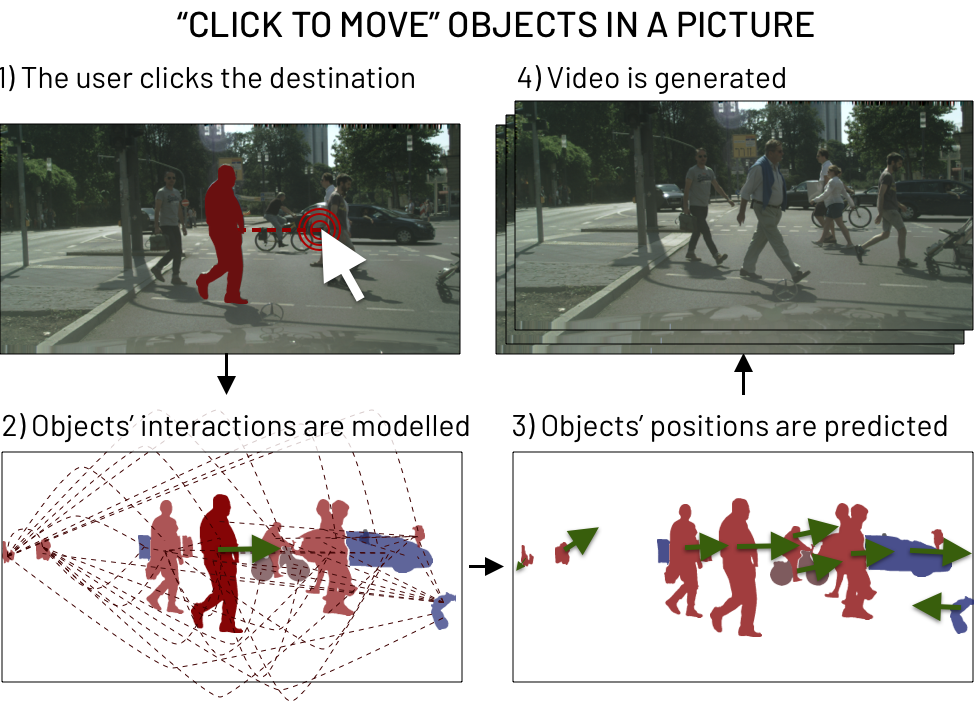}
	\caption{Illustration of the video generation process of Click to Move (C2M): 1) the user selects the objects in a scene and specify their movements. 2) Our network models the interactions between \emph{all} objects through the GCN and 3) predicts their displacement. 4) The network produces a realistic and temporally consistent video.}
	\label{fig:teaser}
\end{figure}

This paper introduces Click to Move (C2M), the first approach that allows users to generate videos in complex scenes by conditioning the movements of specific objects through mouse clicks. Fig.\ref{fig:teaser} illustrates the video generation process of C2M.  The user only needs to select few objects in the scene and to specify the 2D location where each object should move. Our proposed framework receives as inputs an initial frame with its segmentation map and synthesizes a video sequence depicting objects for which movements are coherent with the user inputs.
The proposed deep architecture comprises three main modules: (i) an appearance encoder that extracts the feature representation from the first frame and the associated segmentation map, (ii) a motion module that predicts motion information from user inputs and image features, and (iii) a generation module that outputs the synthesised frame sequence. 
In complex scenes with multiple objects, modelling interactions is essential to generate coherent videos. To this aim, we propose to adopt a Graph Neural Network (GCN), which models object interactions and infers the plausible displacements for all the objects in the video, while respecting the user's constraints.
Experimental results show that our approach outperforms previous video generation methods on two publicly available datasets and demonstrate the effectiveness of the proposed GCN framework in modelling object interactions in complex scenes. 

Our work is inspired by previous literature that generates videos from an initial frame and the associated segmentation maps \cite{pan2019video,sheng2020high}. From these works, we inherit a two-stage procedure where we first estimate the optical flows between an initial frame and all the generated frames, and subsequently refine the image obtained by warping the initial frame according to the estimated optical flows. However, our framework improves over these previous works as it allows the user the possibility to directly control the video generation process with simple mouse clicks. 
Similarly to the work of Hao \emph{et al.}~\cite{hao2018controllable}, we propose to control object movements via sparse motion inputs. However, thanks to the GCN, our approach can deal with scenes with multiple objects, while \cite{hao2018controllable} cannot. 
Furthermore, the method in \cite{hao2018controllable} does not explicitly consider the notion of \emph{object}, as it does not use any instance segmentation information, and does not model the temporal relation between multiple frames. We instead work on multiple frames and in the semantic space, so the user can intuitively select the object of interest and move it in a temporal consistent way. The use of semantic information is motivated by recent findings in the area of image manipulation where it has been shown that semantic maps are beneficial in complex scenes~\cite{ardino2020semantic, lee2018context}.

\textbf{Contributions.} Overall, the main contributions of our work are as follows: 
\begin{itemize} \vspace{-0.6\baselineskip}
\item We propose Click to Move (C2M), a novel approach for video generation of complex scenes that permits user interaction by selecting objects in the scene and specifying their final location through mouse clicks. \vspace{-0.6\baselineskip}
\item We introduce a novel deep architecture that leverages the initial video frame and its associated segmentation map to compute the motion representations that enable the generation of frame sequence. Our deep network incorporates a novel GCN that models the interaction between objects to infer the motion of all the objects in the scene. \vspace{-0.6\baselineskip}
\item Through an extensive experimental evaluation, we demonstrate that the proposed approach outperforms its competitors \cite{pan2019video,sheng2020high} in term of video quality metrics and can synthesize videos where object movements follow the user inputs. \vspace{-0.6\baselineskip}
\end{itemize}

\section{Related Works}

\noindent\textbf{Video generation with user control.} With the recent progress in deep video synthesis, researchers have focused in designing new approaches that include user input in the generation process. Video generation can be controlled by different means. For example, MoCoGAN~\cite{tulyakov2018mocogan} disentangles videos into motion and content latent spaces. Therefore, it is possible to control videos by ``copying" the action from another video or by changing the identity of the person. 
Chan \emph{et al.}~\cite{chan2019everybody} propose to generate dance videos following a ``do as I do" motion transfer strategy: body poses are estimated for every frame of another video and transferred to control the pose of the person in the generated video. Wiles \emph{et al.}~\cite{wiles2018x2face} control human face motion through a driving vector that can be extracted from videos or pose information. Siarohin \emph{et al.}~\cite{siarohin2019animating, siarohin2020first} propose an approach suitable to arbitrary objects and learn motion representations without requiring specific prior knowledge. This approach can be employed with various types of videos, ranging from human bodies to robotics.
Regarding audio-visual methods, talking heads video can be generated from an initial image and an input audio clip~\cite{wiles2018x2face,chen2019hierarchical, zhou2020makelttalk}.
In this paper, we propose a novel framework that involves the user in the generation process. 
However, while previous works mostly focus on generating videos depicting a single object (e.g. a face or human body), we address the more challenging task of video synthesis of complex scenes where multiple objects have to move consistently while accounting for user input.

\noindent\textbf{Future frame prediction.} The problem we address in this work is closely related to future frame prediction, which aims to generate a video sequence given its initial frames. 
Early works formulate the problem as a deterministic prediction task \cite{finn2016cdna,mathieu2015deep,vondrick2015anticipating}. However, this formulation cannot work on most real world videos due to the inherent motion uncertainty.
Thus, recent approaches adopt adversarial~\cite{kumar2019videoflow} or variational~\cite{franceschi2020stochastic,lee2018savp,tulyakov2018mocogan} formulations that can model stochasticity.
Several works focus on the architectural design and propose to estimate optical flow \cite{kumar2019videoflow,finn2016unsupervised, liang2017dual,li2018flow} to generate the future frames by warping the previous one. Others works study solutions for long term predictions ~\cite{villegas2017learning, ho2019sme, ye2019compositional, reda2018sdc}. Similarly, Li \emph{et al.}~\cite{li2018flow} propose a multi-step network that first generates an optical flow, then converts it back to the RGB space to generate novel videos. Instead, Zhang \emph{et al.}~\cite{zhang2020dtvnet} propose to employ an optical flow encoder that maps motion information to a latent space. At test time, different random motion vectors can be sampled to generate video with different motion.

When it comes to complex environment involving multiple objects, additional supervision is highly beneficial. For example, Wu \emph{et al.}~\cite{wu2020future} use video frames, optical flows, instance maps and semantic information together to decouple the background from the dynamic objects and thus predict their trajectory. Similarly, Hao \etal~\cite{hao2018controllable} show that providing sparse motion trajectories to their model helps generating videos with higher quality. However, contrary to our approach, their method does not take advantage of instance segmentation and does not model object interactions.

Recently, Pan \emph{et al.}~\cite{pan2019video} and Sheng \emph{et al.}~\cite{sheng2020high} have proposed to get a benefit from segmentation information to improve video generation. Videos are generated from a single frame and the corresponding segmentation map. Both approaches are based on a two-stage procedure. The first stage aims at estimating the optical flow between the initial frame and every generated frame. In the the second stage, the initial frame is warped according to the optical flow and refined by an encoder-decoder network.  
Inspired from these works, our approach adopts a similar variational auto-encoder framework boosted with optical flow and occlusion supervision. However, we include a novel Graph Convolutional Network (GCN) that models object interactions and takes into account the sparse motion vectors provided by the user.

\begin{figure*}[ht]
    \centering
    \includegraphics[width=0.99\textwidth]{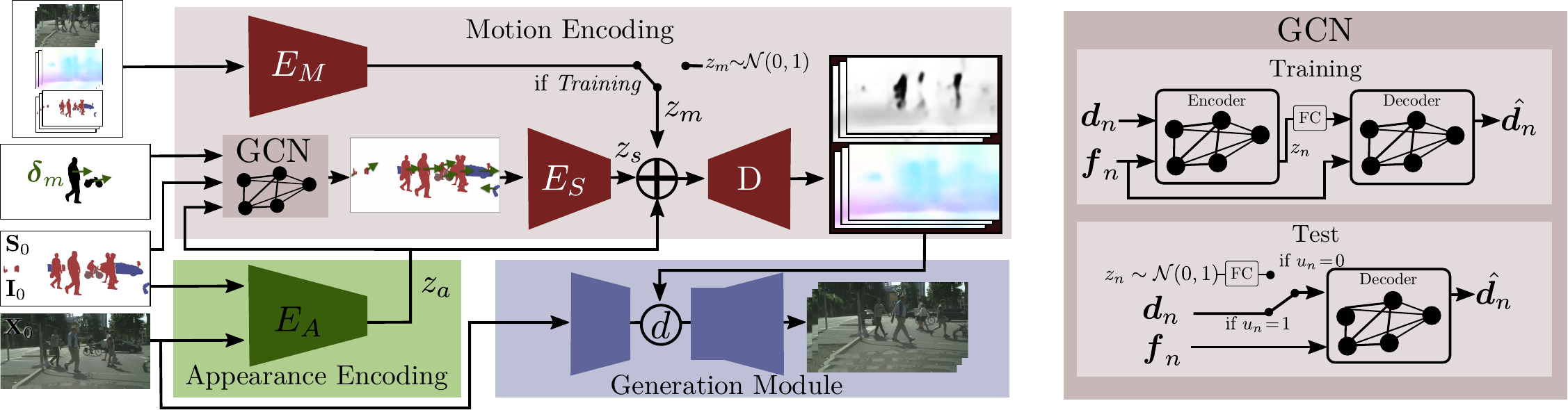}
    \caption{Our network is composed of three modules, namely (i) Appearance encoding, (ii) Motion encoding, and (iii) Generation module. The Appearance Encoding focuses on learning the visual appearance from $\Xmat_0$. The Motion Encoding models the interactions between the objects, predicts their displacement, encodes the motion, and generates the optical flow and occlusion mask for the Generation Module, which focuses on generating temporal consistent and realistic videos. On the right, we show our GCN module to model objects' interactions.}
    \label{fig:pipeline}
\end{figure*}

\section{Click to Move framework}
We aim at generating a video from its initial frame $\Xmat_0\in\mathbb{R}^{H\times W\times 3}$ and a set of user-provided 2D vectors that specify the motion of the key objects in the scene. At test time, we assume that we also have at our disposal the instance segmentation maps of the initial frame. 
Our system is trained on a dataset of videos composed of $T$ frames with the corresponding instance segmentation maps at every frame. As we will see later, in practice, instance segmentation is obtained using a pre-trained model.

Considering a  set of $C$ classes, we assume that $N$ objects are detected at time $t$ in the frame $\Xmat_t\in\mathbb{R}^{H\times W\times 3}$. The instance segmentation is represented via a segmentation map $\Smat_t\in\{0,1\}^{H\times W\times C}$, a class label map $\Cmat_t\in\{1,...,C\}^{H\times W}$ and an instance map $\Imat_t\in\{1,...,N\}^{H\times W}$ that specifies the instance index for every pixel.
At test time, the user provides the motion of the $M$ objects in the scene by drawing 2D arrows corresponding to the displacement between the barycenter of the object in $\Xmat_0$ and the object's desired position at time $T$ (See Fig.~\ref{fig:teaser}). 
Notably, the user is free to provide motion vectors for as many objects as desired. 
Therefore the motion vectors are represented by a list $\mathcal{M}=\{(\deltavect_m,i_m), 1\leq m\leq M\}$, where $\deltavect_m\in\mathbb{R}^2$ contains the barycenter displacement of the object with instance index $i_m$. At training time, the list $\mathcal{M}$ is obtained by randomly sampling objects in every video and estimating their corresponding $\deltavect_m$, which is defined as the displacement of the instance segmentation's barycenters between the first and last frame.

The proposed framework is articulated in three main modules, as illustrated in Fig.~\ref{fig:pipeline}. First, the \emph{Appearance encoding} is in charge of encoding the initial frame. This module receives as input the concatenation of the initial frame $\Xmat_0$, the segmentation $\Smat_0$ and the instance map $\Imat_0$, while it outputs a feature map $\zvect_a$ via the use of an Encoder $E_A$.
Second, the \emph{Motion encoding}, predicts the video motion from the motion vectors provided by the user and the image features $\zvect_a$. This module includes a novel Graph Convolutional Network (GCN) that infers the motion of all the objects in the scene by combining the object motion vectors in $\mathcal{M}$ and the image features $\zvect_a$. This motion module is described in Sec.~\ref{sec:motion} while the details specific to our GCN are given in Sec.~\ref{sec:GCN-update}.
Finally, the \emph{Generation module} is in charge of combining the encoded appearance and the predicted motion to generate every frame of the output video.

\subsection{Object motion estimation with GCNs}
Our GCN aims at inferring the motion of all the objects in the scene by combining the motion vectors provided by the user and the image features $\zvect_a$. 
This section first describes the specific message-passing algorithm that we introduce to model the motion vectors. Then we show how our GCN is embedded into a Variational Auto-Encoder (VAE) framework to allow sampling the possible object motions that respect the user's constraints.

\noindent\textbf{Handling user control with GCNs.}\label{sec:GCN-update} We propose to use a graph to model the interactions between the objects in the scene. Each node corresponds to one of the $N$ objects detected in $\Xmat_0$. The graph is obtained fully connecting all the objects with each other. Let us introduce the following notations: $\fvect_n$ is the feature vector for the $n^{th}$ object and is extracted from $\zvect_a$ via region-wise average pooling. $\dvect_n\in\mathbb{R}^2$ is the estimated barycenter displacement for the $n^{th}$ object. Finally, $u_n\in\{0,1\}$ is a binary value that specifies whether the object motion has been provided by the user ($u_n\!=\!1$) or if it should be inferred ($u_n\!=\!0$). 

In a standard GCN \cite{wu2020comprehensive}, the layer-wise propagation rule specifies how the features $\fvect_n^{(k)}$ at iteration $k$ of the node $n$ are computed from the features of it neighbouring nodes at the previous iteration $\fvect_j^{(k-1)}$:
\begin{equation}
\fvect_n^{(k)}=\sum_{j\in\mathbf{N}(n)}\frac{1}{\sqrt{\mathcal{D}_{nj}}}\thetavect^\top \fvect_n^{(k-1)} \label{eq:GCNupdate} 
  \end{equation}
where $\mathbf{N}(n)$ denotes the neighbours of the node $n$, $\thetavect$ are the trainable parameters and $\mathcal{D}_{nj}$ is a normalization factor equal to the sum of the degree of the nodes $n$ and $j$. In our context, we need to modify this update rule to take into account that the object motion of each node is either known or unknown.    
Besides, we propose two different propagation rules for the node features $\fvect_n$ and the motion vectors $\dvect_n$. We propose to make these rules depending on $u_n$. If $u_n=1$, the node corresponds to an object with a motion controlled by the user and we update only the features: 
\begin{align}
  \fvect_n^{(k)}&=\fvect_n^{(k-1)}\! +\!\sum_{j\in\mathbf{N}(n)}\frac{1}{\sqrt{\mathcal{D}_{nj}}}\thetavect_f^\top (\fvect_n^{(k-1)}\oplus \dvect_n^{(k-1)}) \label{eq:updateun1} \\
  \dvect_n^{(k)}&=\dvect_n^{(k-1)}.
\end{align}
Here, $\thetavect_f$ denotes the trainable parameters and $\oplus$ is the concatenation operation.
This formulation allows propagating feature information through the node while keeping the object motion constant for the nodes with known motion. 
Note that, in \eqref{eq:updateun1}, we opt for a residual update since the messages from the neighbouring nodes are added to the current value $\fvect_n^{(k-1)}$. 
Our preliminary results showed that \eqref{eq:GCNupdate} update rule ended up with all the nodes having the exact same features. 
On the contrary, the residual update that helped objects converging to better features. 
Indeed, this residual update can be seen as skip connections, similar to those of resnet architectures, that allow gradient information to pass through the GCN updates and mitigate vanishing gradient problems.

If $u_n\! = \!0$, the node corresponds to an object with unknown motion and we update both the features and the motion vector. The feature update remains identical to \eqref{eq:updateun1} and the motion vector is updated as follows:
\begin{align}
  \dvect_n^{(k)}&=\dvect_n^{(k-1)}\! +\! \sum_{j\in\mathbf{N}(n)}\frac{1}{\sqrt{\mathcal{D}_{nj}}}\thetavect_d^\top (\fvect_n^{(k-1)}\oplus \dvect_n^{(k-1)}) \label{eq:updateun0}
\end{align}
where $\thetavect_d$ denotes the trainable parameters for the motion estimation. This novel propagation rule allows to aggregate the information contained in the neighbouring nodes to refine the motion estimation of nodes with unknown motion. In the next section, we detail how this GCN is embedded into a VAE framework in order to sample possible object motions.

\noindent\textbf{Overall architecture for motion sampling.}
Our GCN is embedded into a VAE framework composed of an encoder and a decoder network. At training time, we employ an encoder and a decoder while only the decoder is used at test time, as illustrated in Fig.~\ref{fig:pipeline}-Right. Note that the features $\fvect_n$ condition both the encoder and the decoder. The goal of the encoder network is not map the input value $\dvect_n$ of every node to a latent space $\zvect_n$. This encoder is implemented using a GCN that employs the propagation rule described in Sec~\ref{sec:GCN-update} and receives as input $\fvect_n\oplus \dvect_n$ for every node. For every node, the latent variable $\zvect_n$ is given by $\fvect_n^{(k)}$ after the last message propagation update. We assume $\zvect_n$ follows a unit Gaussian distribution ($\zvect_n\! \sim\! \mathcal{N}(0,1)$). The decoder network receives as input the randomly sampled latent variable $\zvect_n$ for the nodes with unknown motion (\ie $u_n=0$) and is trained to reconstruct the input motion $\dvect_n$. 
The decoder is implemented with another GCN with the same propagation rules and with inputs $\fvect_n^{(0)}\oplus \dvect_n^{(0)}$ where $\fvect_n^{(0)}=\fvect_n$ and:
\begin{equation}
    \dvect_n^{(0)}=\begin{cases}
      \text{FC}(\zvect_n) & \text{if $u_n=0$}\\
      \displaystyle\sum_{m=1}^M\mathbbm{1}(i_m=n)\deltavect_m  & \text{if $u_n=1$}.\label{eq:inputDecGCN}
    \end{cases}       
\end{equation}
where $\mathbbm{1}$ denotes the indicator function and
$\text{FC}(.)$ denotes a fully-connected layer that projects the sampled latent variable $\zvect_n$ to the space of $\dvect_n$ (\ie $\mathbb{R}^2$). Intuitively, the sum in \eqref{eq:inputDecGCN} iterates over all the objects in $\mathcal{M}$ to select the corresponding motion vector provided by the user.

At test time, the GCN encoder is not used. The latent variable $\zvect_n$ is sampled according to our unit Gaussian prior distribution for every object with unknown motion and forwarded to the decoder. The decoder outputs the 2D motion of every object in the scene.

\subsection{Motion encoding}
\label{sec:motion}
This module is in charge of predicting the optical flows and the occlusion maps between the initial frame $\Xmat_0$ and every frame that has to be generated. To this aim, for every time step $t$, we compute a binary tensor $\Bmat_t\in\{0,1\}^{H\times W}$ that specifies the locations of the objects in the scene. 
At time $t\! =\! 0$, the object-location map $\Bmat_0$ is computed from the instance segmentation map $\Imat_0$:%
\begin{equation}%
  \forall (i,j)\in H\times W, \Bmat_0[i,j]=\sum_n^N \mathbbm{1}(\Imat_0[i,j]=n).\label{eq:Omat}%
\end{equation}
For $t>0$, $\Bmat_t$ cannot be estimated with the previous equation since $\Imat_t$ is not known at test time. Instead, we consider a simple rigid model for every object and obtain $\Bmat_t$ by warping $\Bmat_0$ according to the the object motion $\dvect_t$. At training time, $\dvect_t$ is estimated from the segmentation maps while, at test time, we employ $\hat{\dvect}_t$, which is the displacement predicted by our GCN. Finally, this object-location tensor is mapped to a latent tensor $\zvect_s$ via an encoder $E_S$. 

Note that, the output video cannot be fully encoded via the initial frame and the motion of each object since there exist other sources of variability such as the appearance of new objects or change in object sizes. Therefore, we introduce a latent motion variable $\zvect_m$ that encodes all the motion information that cannot be described by $\zvect_s$ and $\zvect_a$. We employ an auto-encoder strategy at training time, estimating $\zvect_m$ from the complete video sequence with an encoder $E_M$. More precisely, $E_M$ receives as input the concatenation of all the video frames, the instance segmentation maps $S_0$ and $I_0$, and the optical flow for every frame. At test time, the latent motion code $\zvect_m$ is sampled according to the prior distribution (\ie $\zvect_m\sim \mathcal{N}(0,I)$). 

Finally, we provide the latent variables $\zvect_a$, $\zvect_s$ and $\zvect_m$ to the same decoder, which outputs the bi-directional optical flows and occlusion maps. More precisely, the decoder outputs the forward and the backward optical flow at every time steps denoted by $\Fmat_t^f$ and $\Fmat_t^b$ respectively and the corresponding occlusion maps  $\Omat_t^f$ and $\Omat_t^b$. Note that the backward optical flows and occlusion maps are then provided to the generation modules, while the forward optical flow and occlusion maps are used only for loss computation.

\subsection{Generation module and training objectives}
We employ a generation module inspired by \cite{siarohin2020first}. After two down-sampling convolutional blocks applied on the initial frame $\Xmat_0$, we obtain a feature map. We proceed independently for every frame to generate and warp the feature map according to the optical flow predicted by the motion module. Then we multiply the warped feature map by the occlusion map predicted by the occlusion estimator to diminish the impact of the features corresponding to the occluded parts. Finally, the masked feature maps are fed to a subsequent network to output the generated video. This network is composed of several residual blocks, followed by two up-sampling convolutional blocks. 

\noindent\textbf{Objective functions.}
Our GCN framework employs the evidence lower bound of the VAE framework. It is composed of a reconstruction term on the predicted 
motion vector and the Kullback-Leibler divergence (KL) between the conditional distribution of $\zvect_n$ and its unit Gaussian prior:  
\begin{equation}
    \mathcal{L}_{VAE} = \frac{1}{N} \sum_{n=0}^N \| \dvect_n - \hat{\dvect}_n \|_1 - \mathcal{D}_{KL}(\zvect_n \| \mathcal{N}(0,I)),
\end{equation}
where $\hat{\dvect}_n$ is the displacement predicted by the GCN.

\noindent\textit{Forward-backward Consistency.}
Similarly to~\cite{shen2020interpreting}, we ensure the cycle consistency between forward and backward optical flows.  More precisely, for every non-occluded pixel location $\pvect$, we minimize the $L_1$ distance between the corresponding optical flows:  
\begin{equation}
\begin{aligned}
\mathcal{L}_{Fc}(F^f, F^b) =  \frac{1}{T}  \sum_{i=1}^T &\sum_{{\pvect}} \Omat_t^f({\pvect}) |\Fmat^f_t({\pvect})-{\Fmat}^b_t({\pvect}+{\Fmat}^f_t({\pvect}))|_1 \\
 & + {\Omat}_t^b({\pvect}) |{\Fmat}^b_t({\pvect})-{\Fmat}^f_t({\pvect}+{\Fmat}^b_t({\pvect}))|_1\label{eq:lossOF}
\end{aligned}
\end{equation}

\noindent\textit{Smoothness.} Following~\cite{sheng2020high}, we employ a smoothness loss that penalizes high gradient values in the optical-flow map that do not correspond to high-gradient values in the image $\Xmat_0$ (for more details refer to ~\cite{sheng2020high}).

\noindent\textit{Supervised flow.} To improve the quality of the generated videos in our multi-objects setting, we take advantage of a pre-trained FlowNet2~\cite{ilg2017flownet} network for optical flow and occlusion estimation. FlowNet2 provides high quality optical flow maps that we use as supervision for our motion decoder network using a standard L1 loss.

\noindent\textit{Motion Encoding uncertainty.}
To allow the sampling of $\zvect_m$ at test time, the output of the motion encoder $E_M$ is mapped to a unit Gaussian distribution via the KL-divergence:
\begin{equation}
\begin{aligned}
\mathcal{L}_{m} =& - \mathcal{D}_{KL}(z_m \| \mathcal{N}(0,I))
\end{aligned}
\end{equation}

\noindent\textit{Generation module.} 
The generation module is trained using state-of-the-art losses for video generation. Following~\cite{mao2017least, wang2018high, isola2017image} we adopt a PatchGAN discriminator trained with a Least Square loss.
For the generator, we apply the structural similarity loss~\cite{wang2004image}, the perceptual loss~\cite{johnson2016perceptual}, feature matching loss~\cite{wang2018high}, and a standard pixel-level reconstruction L1 loss.

\section{Experiments}

\noindent\textbf{Datasets.} We evaluate our model with two publicly available datasets, namely Cityscapes and KITTI 360.
\begin{itemize}[leftmargin=*,noitemsep,topsep=0pt]	
    \item \emph{Cityscapes}~\cite{cordts2016cityscapes} provides videos at 17 Frames Per Second (FPS) of European urban scenes. We resize all images to $256 \times 128$ resolution for performance reasons. The dataset contains 2975 video sequences for training and 500 video sequences for testing. Since Cityscapes does not provides instance and semantic segmentations for the video sequences, we used~\cite{cheng2020panoptic} to generate them.
    \item \emph{KITTI 360}~\cite{Xie2016CVPR} provides a richly annotated videos at 11 FPS in German suburban areas. We resize all images to $192 \times 64$ resolution. The dataset for our evaluation contains 6941 training videos and 423 test sequences. We aggregate the segmentation categories to match the 19 classes of Cityscapes.
\end{itemize}

\noindent\textbf{Baselines.} We compare with the state-of-the-art model for video generation in complex scenarios, \textit{i.e.} Sheng \emph{et al.}~\cite{sheng2020high}, which can generate high-quality videos from a staring frame and its associated semantic segmentation map. Since Sheng \emph{et al.}~\cite{sheng2020high} is not able to generate videos controlling object positions, we modify it by including the object location tensor $\Bmat_t$ into the appearance encoder of the original model. We call this model \emph{Sheng*}. For a fair comparison, we also test our approach with a variant of the method of Sheng \emph{et al.}, referred to as \emph{S. Sheng*}, where we add our \emph{Supervised flow} loss that uses the supervision of a pretrained network in order to improve optical flow prediction.
We note that Sheng \emph{et al.}~\cite{sheng2020high} is an extension of Pan \emph{et al.}~\cite{pan2019video} and that these two works correspond to the same method. Thus, Pan \emph{et al.}~\cite{pan2019video} is not included in our comparison. 
It is also worth that Hao \emph{et al.}~\cite{hao2018controllable} is not included in the baselines, as it focuses on image generation and does not \emph{explicitly} model the semantic space. Thus, it would be unfair to compare Hao \emph{et al.} with our method on the temporal consistency and object displacements in videos.

\noindent\textbf{Settings.} We design three test settings to evaluate our proposal extensively.
\begin{itemize}[leftmargin=*,noitemsep,topsep=0pt]	
    \item \emph{Oracle (O)}. For each video, we select a random object that has to be moved, we feed the networks with the ground truth displacements between the first and last frames, and let the models generate the video. This setting evaluates the network capacity to benefit from the given sparse motion information.
    \item \emph{Custom}. For each input video, we select a random object that has to be moved, we feed the networks with displacement shifted by $\lambda=1.5$ (i.e. $\dvect_n' = \lambda \dvect_n $) and let the models generate the video. This setting evaluates the network capacity to condition the video on sparse motion inputs, which are different from the ground truth.
\end{itemize}
Then, we also experiment a drastic scenario where \emph{all} the objects are moved following the \emph{Custom}. In this experiments, all future positions are provided as input. In this experiment, the GCN can be by-passed since $u_n=1$ for every object. This experiments differ from \emph{Ground truth} and \emph{Custom} where our GCN has to infer the plausible future positions of all the objects that are not provided by the user.
In all our experiments, we generate 5 future frames starting from the provided initial frame.

\noindent\textbf{Evaluation metrics.}
\begin{itemize}[leftmargin=*,noitemsep,topsep=0pt]	
\item \emph{FVD.} We adopt the Fréchet video distance (FVD) metric~\cite{unterthiner2019fvd} to evaluate both the video quality and temporal consistency of generated frames. We compute the FVD between the ground truth test videos and the generated ones. The lower the FVD, the better.

\item \emph{NDE.} We measure the adherence of generated videos with the user-provided motions by computing the Normalised Displacement Error (NDE) as the Euclidean distance between the coordinate specified by the user and the coordinate where the object ends-up in the generated video, which is then normalised Euclidean distance of the ground truth starting coordinate and the ending one.
All object's positions are detected through YOLOv3 ~\cite{redmon2018yolov3}. We discard the objects that cannot be detected in the ground truth videos due to the resolution of videos, or because objects are too small to be correctly detected by YOLOv3. 
The lower the NDE, the better.

\item \emph{Acc.} The object's positions in generated videos can be difficult to track due to the presence of artifacts, occlusions and low-quality images. Thus, we report here the Accuracy (Acc) of the YOLOv3 detector in generated videos. The higher the Accuracy, the better.
\end{itemize}

\begin{table}[!h]
    \centering
    \begin{tabular}{@{}l rrr@{}}
	\toprule
		\textbf{Model} & \textbf{FVD$\downarrow$} & \textbf{NDE$\downarrow$} & \textbf{Acc$\uparrow$}\\
		\midrule
		A: Our proposal& 288 & 1.01 & 0.84 \\
	    B: (A) w/o GCN & 369 & 1.42 & 0.70\\
	    C: (A) w/o Obj. Interactions & 375 & 1.38 & 0.76\\
		D: (A) w/o Sup. & 301 & 1.13 & 0.84\\
    \bottomrule      
    \end{tabular}
    \caption{Ablation study results on Cityscapes.}
    \label{tab:ablation}
    \vspace{-2mm}
\end{table}

\begin{table}[ht]
\centering
\small
    \setlength{\tabcolsep}{3pt}
    \begin{tabular}{@{}ll rrr rrr@{}}
    \toprule
    \multirow{2}{*}{\specialcell{\textbf{Setting}\\($N$)}} & \multirow{2}{*}{\textbf{Model}} &
    \multicolumn{3}{c}{\textbf{Cityscapes}} &  \multicolumn{3}{c}{\textbf{KITTI 360}} \\
    \cmidrule(l{2pt}){3-5} \cmidrule(l{2pt}){6-8}
    && FVD$\downarrow$ & NDE$\downarrow$ & Acc$\uparrow$ & FVD$\downarrow$ & NDE$\downarrow$ & Acc$\uparrow$ \\
    \midrule
    \multirow{4}{*}{\specialcell{\emph{Oracle}\\(1)}} & Sheng~\cite{sheng2020high} & 373 & 2.11 & 0.68 & \textbf{443} & 3.92 & 0.68\\
     & Sheng* & 498 & 2.12 & 0.58  & 507 & 3.66 & 0.66 \\
     & S. Sheng* & 493 & 1.78 & 0.57 & 527 & 3.79 & 0.33\\
     & Ours  & \textbf{288} & \textbf{1.01} & \textbf{0.84} & 463 & \textbf{1.83} & \textbf{0.75}\\
    \midrule \midrule
    \multirow{4}{*}{\specialcell{\emph{Custom}\\(1)}} & Sheng~\cite{sheng2020high} & 373 & 1.53 & 0.66 & \textbf{443} & 3.98 & 0.62 \\
     & Sheng* & 498 & 1.61 & 0.57 & 506 & 3.27 & 0.60 \\
     & S. Sheng* & 493 & 1.41 & 0.59 & 527 & 3.34 & 0.30 \\
     & Ours & \textbf{303} & \textbf{0.66} & \textbf{0.88} & 470 & \textbf{2.06} & \textbf{0.81}\\
    \midrule \midrule
    \multirow{4}{*}{\specialcell{\emph{Custom}\\(all)}} & Sheng~\cite{sheng2020high} & 373 & 1.48 & 0.73 & \textbf{443} & 2.93 & 0.48\\
     & Sheng* & 498 & 1.47 & 0.67 & 506 & 3.19 & 0.49\\
     & S. Sheng* & 493 & 1.38 & 0.60 & 527 & 2.71 & 0.24 \\
     & Ours & \textbf{321}  & \textbf{0.96} & \textbf{0.86} & 464& \textbf{1.58} &  \textbf{0.72}\\
    \bottomrule    
    \end{tabular}%
    \caption{Quantitative comparison in the \emph{Oracle}  and \emph{Custom} setting. $N$ is the number of user-controlled objects. $N=1$ selects one object at random}
    \label{tab:results}
\end{table}

\vspace{-3mm}
\subsection{Ablation Study}
We conduct an ablation study on Cityscapes to evaluate the impact of the individual components of the model. 
We begin by testing the contribution of our GCN by removing the motion estimation module and directly use the object location tensor of the user-controlled object $\Bmat_t$ in the appearance encoder. \Cref{tab:ablation}-B shows that removing the motion estimator leads to a drop in all three metrics. Without the GCN, the network cannot infer the positions of the objects in the scene and fails at moving the object. The quality of the video decreases as well (FVD 369 vs FVD 289). 

Then, we test a version of the GCNs that does not model the interactions between objects. To do so, we remove all the edges between the nodes of the GCN, thus considering each object as independent. Tab.~\ref{tab:ablation}-C shows that, while the object is correctly moved (NDE and Acc are similar to A), the video quality is considerably worse. In the Supplementary Material, we qualitatively show that the network cannot move the other objects consistently.

Finally, we also test the network without flow supervision (\ie Tab.~\ref{tab:ablation}-D). As expected, the performance decreases in NDE and FVD. Nevertheless, the quality of the image quality measured with FVD remains higher than when we do not model object interactions (\ie Tab.~\ref{tab:ablation}-C).

\begin{figure}[t]
    \centering
  \setlength{\tabcolsep}{1pt}
	\renewcommand{\arraystretch}{0.8}
    \newcommand{\sizea}{0.31\linewidth}
    \footnotesize
	\begin{tabular}{cccc}
	  &\emph{t} +  1 & \emph{t} +  3 & \emph{t} +  5 \\
      \rotatebox{90}{\ \ \cite{sheng2020high}} &
      \includegraphics[width=\sizea]{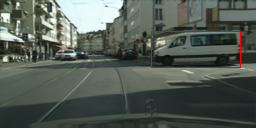} & \includegraphics[width=\sizea]{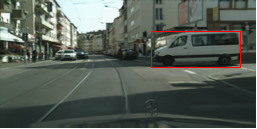} & \includegraphics[width=\sizea]{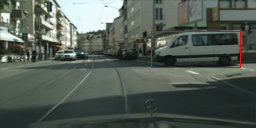}
      \\
	  \rotatebox{90}{\ \ $\lambda$=$1$} &
	  \includegraphics[width=\sizea]{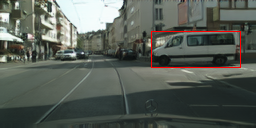}&
	  \includegraphics[width=\sizea]{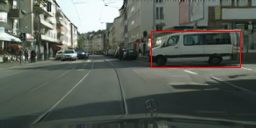} & \includegraphics[width=\sizea]{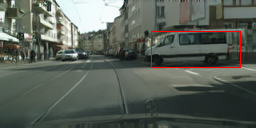}\\
	   \rotatebox{90}{$\lambda$=$1.5$} &
	   \includegraphics[width=\sizea]{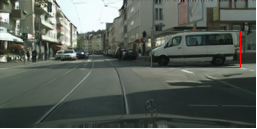} & \includegraphics[width=\sizea]{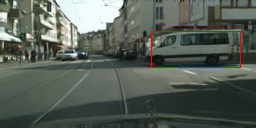} & \includegraphics[width=\sizea]{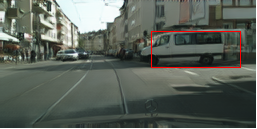}\\
	   \rotatebox{90}{\ \ \ \ \ GT} &
	   \includegraphics[width=\sizea]{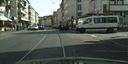} & \includegraphics[width=\sizea]{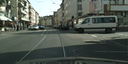} & \includegraphics[width=\sizea]{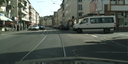} \\
	   
	\end{tabular}%
	\caption{Qualitative comparison in the \emph{Custom} setting on the Cityscapes dataset with ground truth reference. The position of the moved object at $t=0$ is highlighted in red. Zoom for details.}%
\label{Fig:Van}%
\end{figure}

\begin{figure*}[ht]
    \setlength{\tabcolsep}{1pt}
	\renewcommand{\arraystretch}{0.8}
    \newcommand{\sizea}{0.16\linewidth}
    \footnotesize
	\centering
	\begin{tabular}{ccccccc}
	  &\emph{t} +  1 & \emph{t} +  3 & \emph{t} +  5 & \emph{t} +  1 & \emph{t} +  3 & \emph{t} +  5 \\
      \rotatebox{90}{\ Sheng \cite{sheng2020high}} & \includegraphics[width=\sizea]{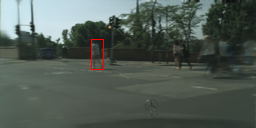} & \includegraphics[width=\sizea]{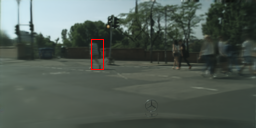} & \includegraphics[width=\sizea]{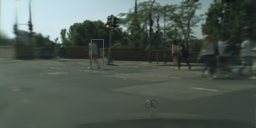} & \includegraphics[width=\sizea]{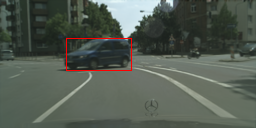} & \includegraphics[width=\sizea]{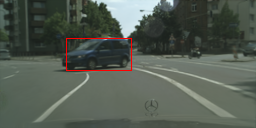} & \includegraphics[width=\sizea]{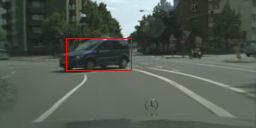}
      \\
	  \rotatebox{90}{\ \ \ \ Sheng*} & \includegraphics[width=\sizea]{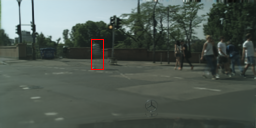} & \includegraphics[width=\sizea]{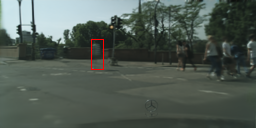} & \includegraphics[width=\sizea]{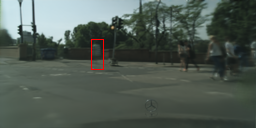} & \includegraphics[width=\sizea]{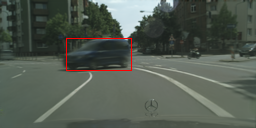}& \includegraphics[width=\sizea]{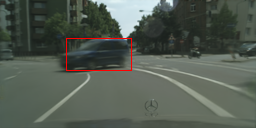}&
	  \includegraphics[width=\sizea]{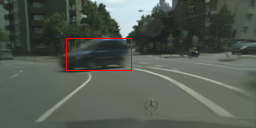}\\
	   \rotatebox{90}{\ \ S. Sheng*}&\includegraphics[width=\sizea]{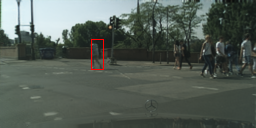} & \includegraphics[width=\sizea]{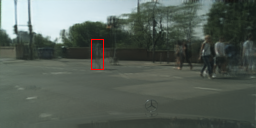} & \includegraphics[width=\sizea]{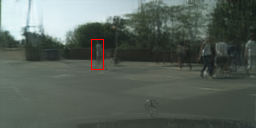} &\includegraphics[width=\sizea]{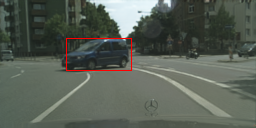}& \includegraphics[width=\sizea]{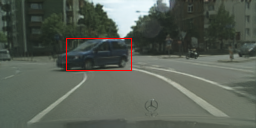}&
	   \includegraphics[width=\sizea]{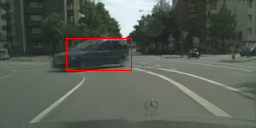}\\
	   \rotatebox{90}{\ \ \ \ \ \ Ours} &\includegraphics[width=\sizea]{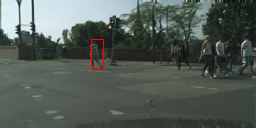} & \includegraphics[width=\sizea]{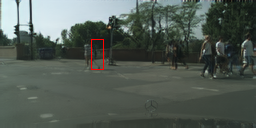} & \includegraphics[width=\sizea]{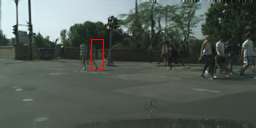} & \includegraphics[width=\sizea]{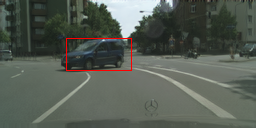} & \includegraphics[width=\sizea]{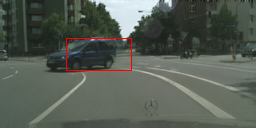} & \includegraphics[width=\sizea]{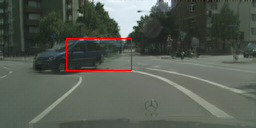}\\
	   	   \rotatebox{90}{\ \ \ \ \ GT} &\includegraphics[width=\sizea]{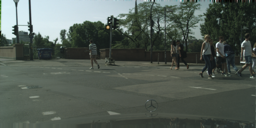} & \includegraphics[width=\sizea]{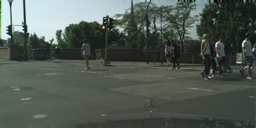} & \includegraphics[width=\sizea]{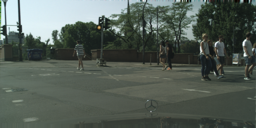} & \includegraphics[width=\sizea]{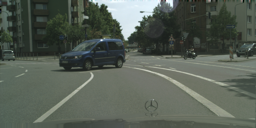} & \includegraphics[width=\sizea]{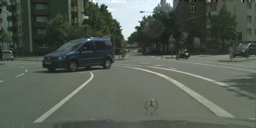} & \includegraphics[width=\sizea]{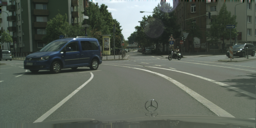}\\
	   
	\end{tabular}
	\caption{Results of predicting the frames \emph{t} + 1,  \emph{t} + 3 , and  \emph{t} + 5 on the Cityscapes dataset~\cite{cordts2016cityscapes} with ground truth reference. On first three columns, we move the pedestrian near the semaphore to left. On the last three columns we move car crossing the street. The position of the moved object at $t=0$ is highlighted in red. Zoom for details.}
	\label{Fig:cityscapes}
\end{figure*}

\subsection{Comparison with State-of-the-art}
\noindent\textbf{Quantitative comparison.} 
We compare our method with the method of Sheng \emph{et al.}~\cite{sheng2020high}, and its modifications, namely \emph{Sheng*} and \emph{S. Sheng*}. To the best of our knowledge, the method of Sheng \emph{et al.}~\cite{sheng2020high} model is the most similar work that generates videos in complex environments also leveraging the semantic space of frames.

Tab.~\ref{tab:results} shows the quantitative evaluation of all the models. We first compare our proposal in the \emph{Oracle} setting, where the displacement $\dvect_n$ of one random object $n$ is computed from ground truth frames. 
From NDE and Acc, we observe that our approach consistently outperforms state-of-the-art methods by enabling the user to move objects more precisely in both the datasets (see Tab.~\ref{tab:results}-Oracle results). In particular, NDE decreases from Sheng \emph{et al.}~\cite{sheng2020high} results by 47\% and 53\% in Cityscapes and KITTI 360, respectively.
Regarding the video quality, which evaluates both the temporal consistency and image quality, we significantly improve the state-of-the-art performance in Cityscapes (FVD decreases by $22.79\%$ from \cite{sheng2020high}), while in KITTI 360 we are slightly worse than Sheng \emph{et al.}~\cite{sheng2020high}.
We hypothesize this result is caused by the low frame rate of KITTI 360, which rewards Sheng \emph{et al.}~\cite{sheng2020high} that is dominated by modelling only the ego-motion, while ignoring other objects' movements. 
Through \emph{Sheng*} results, we note that adding the information to move the objects in the scene helps the baseline through NDE, but the video quality decreases significantly. Only through additional supervision (i.e. \emph{S. Sheng*}), FVD partially decreases. However, our model is far better at moving the objects in the scene, while having better video quality than \emph{Sheng*} and \emph{S. Sheng*}.

Tab.~\ref{tab:results}-\emph{Custom} also shows the \emph{Custom} experiment, where $\dvect_i$ is multiplied by $\lambda=1.5$ from the ground truth displacement. Again, we observe that our proposal offers better control of the object's movements compared to state-of-the-art approaches and improves video quality. Our approach moves objects to positions that differ from the ground truth, showing that the \emph{Motion Encoding} module correctly follows the user inputs, infers the missing objects and composes them in a temporally consistent manner.

We also perform experiments where we ask the models to move \emph{all} objects (i.e. $N$ objects) with the \emph{Custom} setting. The last rows of Tab.~\ref{tab:results} show that our proposal achieves the best results even at this ``drastic" task.

Finally, we note that, our approach without supervised optical flow (Tab.\ref{tab:ablation}-D) outperforms the existing approaches compared in Tab.~\ref{tab:results} both in terms of video quality and object control. This result confirms that the performance of our approach are not due to our use of supervision for optical flow but rather to our architecture.

\noindent\textbf{Qualitative comparison.} 
We now report the qualitative comparison for the tested models. Fig.\ref{Fig:cityscapes} shows the results of two groups of experiments, where we feed the network with two different initial frames. 
In the first group of images, we want to move to the left the pedestrian that in the ground truth is in the position highlighted with the red bounding box.
All three baselines fail to move the pedestrian. Sheng \emph{et al.}~\cite{sheng2020high} only moves the ego vehicle slightly to the right, leaving the pedestrian in the same position, while moving the entire scene. \emph{Sheng*} and \emph{S. Sheng*} moves the ego vehicle forward but fail at moving the pedestrian, which stays exactly in the same position in all the frames.
C2M, instead, correctly and gradually moves to the left of the pedestrian, which goes out the red bounding box. 

In the second group of images in the last three columns of Fig.\ref{Fig:cityscapes}, we aim to move a car that in the ground truth was in the position highlighted in red. Sheng \emph{et al.}~\cite{sheng2020high} can only move the ego-motion forward while the car remains in the same starting position. The other two baselines slightly move the car but not to the desired position specified by the user. However, our proposal significantly moves the car to the left, which goes partially out from the bounding box, while changing very little in the ego-motion of the video.  

Finally, Fig.\ref{Fig:Van} shows a qualitative example of how our model can modify the van's position with different displacement. Moving it to the ground truth position ($\lambda=1$) and to custom coordinates ($\lambda=1.5$). As seen in the previous experiment, the baseline fails at moving the white van to the left. Instead, it stretches the back of the van. In contrast, with $\lambda=1$ and $\lambda=1.5$ the van goes from the bounding box with different horizontal shifts, depicting that our network can correctly change the position of the van to the user-specified positions.

\section{Conclusions}
In this work, we introduce \emph{Click to Move}, a framework for video generation that allows the user to select key objects in the scene and control their motion by specifying their position in the last video frame. At test time, our approach receives the initial frame and the corresponding instance segmentation maps to generate a video that starts from the provided frame and respects the object motion constraints specified by the user. 
Objects in a scene are often not independent one from another. Thus, we introduce a novel GCN framework that employs specific message-passing
rules to model object interaction while accounting for the user inputs. 
Experimentally, we demonstrate that our method outperforms state-of-the-art approaches and that the proposed GCN architecture allows better motion control. As future works, we plan to extend our approach to allow the generation of videos with variable length. 

\section{Acknowledgments}
E. R. acknowledges financial support from the EU project PROTECTOR: Protecting Places of Worship. This work was carried out under the \emph{Vision and Learning joint Laboratory} between FBK and University of Trento.

{\small
\bibliographystyle{ieee_fullname}
\bibliography{egbib}
}

\appendix
\section*{Supplementary material}

We here provide additional information regarding network architectures (\Cref{sec:net}) and implementation details (\Cref{sec:implem}). Then, we provide additional high-resolution qualitative results in \Cref{sec:quali} on both Cityscapes and KITTI 360. 

\section{Network architectures}
\label{sec:net}
We provide the details of the trajectory encoder $E_s$ and flow predictor $D$ in Table~\ref{tab:architecture}. The flow decoder is composed of two heads (referred to as \emph{Flow}$_D$ and \emph{Occ}$_D$) and shared layers (referred to as \emph{Feat}$_D$). For more details regarding the architecture of the other networks, please refer to the code attached with this supplementary material.

\begin{table*}[!ht]
	\centering
	\footnotesize
	\begin{tabular}{@{}lll@{}}
	    \toprule
	    \textbf{Part} & \textbf{Input} $\rightarrow$ \textbf{Output Shape} & \textbf{Layer Information} \\ \midrule
		\multirow{3}{*}{$E_s$} & (BS,5, 64,128, 1 ) $\rightarrow$ (BS,5, 32,64,32) & 3DCONV-(N32, K\{3,4,4\}, S\{1,2,2\}, P\{1,1,1\}), BN, LeakyReLU \\
		    & (BS,5, 32,64,32 ) $\rightarrow$ (BS,5, 16,32,64) & 3DCONV-(N64, K\{3,4,4\}, S\{1,2,2\}, P\{1,1,1\}), BN, LeakyReLU \\
		    & (BS,5, 16,32,64 ) $\rightarrow$ (BS,5, 8,16,128) & 3DCONV-(N128, K\{3,4,4\}, S\{1,2,2\}, P\{1,1,1\}), BN, LeakyReLU \\
		\midrule
		\multirow{13}{*}{$Feat_D$} & (BS * 5, 2, 4, 272) $\rightarrow$ (BS * 5, 2, 4, 512) & CONV-(N512, K3, S1, P1), BN, LeakyReLU \\
		    & (BS * 5, 2, 4, 512) $\rightarrow$ (BS * 5, 4, 8, 256) & UPCONV, CONV-(N256, K3, S2, P1), BN, LeakyReLU \\
		    & (BS * 5, 4, 8, 512) $\rightarrow$ (BS * 5, 8, 16, 128) & SKIP, UPCONV, CONV-(N128, K3, S1, P1), BN, LeakyReLU \\
		    & (BS * 5, 4, 8, 512) $\rightarrow$ (BS, 5, 8, 16, 128) & RESHAPE \\
		    & (BS, 5, 8, 16, 256) $\rightarrow$ (BS, 5, 8, 16, 128) & SKIP, 3DCONV-(N128, K\{3,3,3\}, S\{1,1,1\}, P\{1,1,1\}), BN, LeakyReLU \\
		    & (BS, 5, 8, 16, 128) $\rightarrow$ (BS * 5, 8, 16, 128) & RESHAPE \\
		    & (BS * 5, 8, 16, 256) $\rightarrow$ (BS * 5, 16, 32, 64) & SKIP, UPCONV, CONV-(N64, K3, S1, P1), BN, LeakyReLU \\
		    & (BS * 5, 16, 32, 64) $\rightarrow$ (BS, 5, 16, 32, 64) & RESHAPE \\
		    & (BS, 5, 16, 32, 128) $\rightarrow$ (BS, 5, 16, 32, 64) & SKIP, 3DCONV-(N64, K\{3,3,3\}, S\{1,1,1\}, P\{1,1,1\}), BN, LeakyReLU \\
		    & (BS, 5, 16, 32, 64) $\rightarrow$ (BS * 5, 16, 32, 64) & RESHAPE \\
		    & (BS * 5, 16, 32, 128) $\rightarrow$ (BS * 5, 32, 64, 32) & SKIP, UPCONV, CONV-(N32, K3, S1, P1), BN, LeakyReLU \\
		    & (BS * 5, 32, 64, 32) $\rightarrow$ (BS, 5, 32, 64, 32) & RESHAPE \\
		    & (BS, 5, 32, 64, 64) $\rightarrow$ (BS, 5, 32, 64, 32) & SKIP, 3DCONV-(N32, K\{3,3,3\}, S\{1,1,1\}, P\{1,1,1\}), BN, LeakyReLU \\
		\midrule
		\multirow{2}{*}{$Flow_D$} & (BS * 5, 32, 64, 64) $\rightarrow$ (BS * 5, 64, 128, 32) & SKIP, UPCONV, CONV-(N32, K3, S1, P1), BN, LeakyReLU \\
		    & (BS * 5, 64, 128, 32) $\rightarrow$ (BS * 5, 64, 128, 2) & CONV-(N2, K5, S1, P2), IN, Tanh \\
		\midrule
		\multirow{2}{*}{$Occ_D$} & (BS * 5, 32, 64, 64) $\rightarrow$ (BS * 5, 64, 128, 32) & SKIP, UPCONV, CONV-(N32, K3, S1, P1), BN, LeakyReLU \\
		     & (BS * 5, 64, 128, 32) $\rightarrow$ (BS * 5, 64, 128, 2) & CONV-(N1, K5, S1, P2), Sigmoid \\
		\bottomrule
	\end{tabular}
	\vspace{1mm}
	\caption{Network architecture. We use the following notation: $Z$: the dimension of motion vector, K: kernel size, S: stride size, P: padding size, CONV: a convolutional layer, UPCONV: upsample, 3D-CONV: 3D convolutional layer, BN: Batch Normalization, SKIP: skip connection }
	\label{tab:architecture}
\end{table*}

\subsection{Convergence issue}
During preliminary experiments, we observed that using eq.(1) update rule (from the manuscript) ended up with all the nodes having the exact same features. 
For this reason, we added a residual update that helped objects converging to better features. 
Indeed, this residual update can be seen as skip connections, similar to those of resnet architectures, that allow gradient information to pass through the GCN updates and mitigate vanishing gradient problems.

\section{Implementation.}
\label{sec:implem}
Our architecture is implemented with Pytorch 1.7.0, while the graph neural network has been implemented using Pytorch Geometric. We use the ADAM optimizer~\cite{kingma2014adam} with a learning rate of 2e-4 for the Generation module and 1e-4 for the Motion estimation. The modules are trained upon convergence. Training takes about one day for the Cityscapes dataset and two days for the KITTI 360 dataset. The experiments are done using two Nvidia RTX 2080Ti.

\section{Additional qualitative results}
\label{sec:quali}  
\begin{figure*}[ht]
    \setlength{\tabcolsep}{1pt}
	\renewcommand{\arraystretch}{0.8}
    \newcommand{\sizea}{0.30\linewidth}
    \footnotesize
	\centering
	\begin{tabular}{cccc}
	  &\emph{t} +  1 & \emph{t} +  3 & \emph{t} +  5 \\
      \rotatebox{90}{\ Sheng \cite{sheng2020high}} &       \includegraphics[width=\sizea]{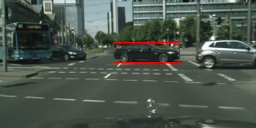} &
      \includegraphics[width=\sizea]{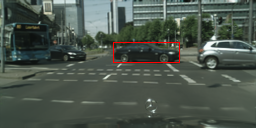} &
      \includegraphics[width=\sizea]{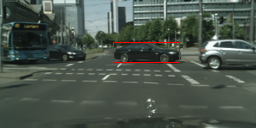}
      \\
	  \rotatebox{90}{\ \ \ \ Sheng*}   &
	  \includegraphics[width=\sizea]{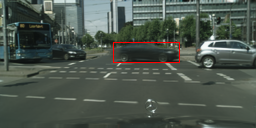} &
      \includegraphics[width=\sizea]{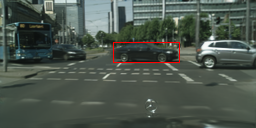} &
      \includegraphics[width=\sizea]{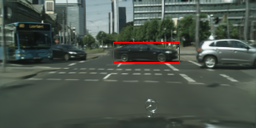}
      \\
	   \rotatebox{90}{\ \ S. Sheng*}&
      \includegraphics[width=\sizea]{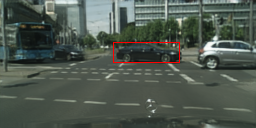} &
      \includegraphics[width=\sizea]{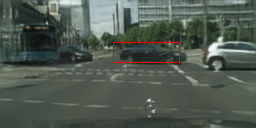} &
      \includegraphics[width=\sizea]{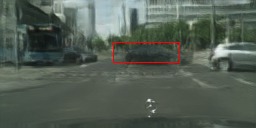}
      \\
	   \rotatebox{90}{\ \ \ \ \ \ Ours} &
      \includegraphics[width=\sizea]{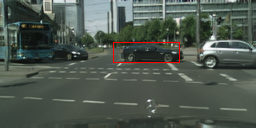} &
      \includegraphics[width=\sizea]{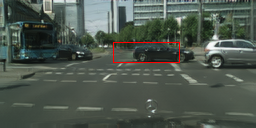} &
      \includegraphics[width=\sizea]{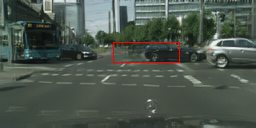}
      \\
	   	   \rotatebox{90}{\ \ \ \ \ GT} &
	  \includegraphics[width=\sizea]{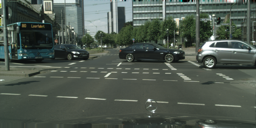} &
      \includegraphics[width=\sizea]{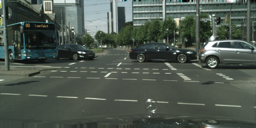} &
      \includegraphics[width=\sizea]{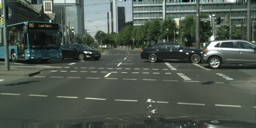}\\
	   
	\end{tabular}
	\caption{Results of predicting the frames \emph{t} + 1,  \emph{t} + 3 , and  \emph{t} + 5 on the Cityscapes dataset~\cite{cordts2016cityscapes} with ground truth reference. On first three columns, we move the pedestrian near the semaphore to left. On the last three columns we move car crossing the street. The position of the moved object at $t=0$ is highlighted in red. Zoom for details.}
	\label{Fig:cityscapes_supplementary1}
\end{figure*}

\begin{figure*}[ht]
    \setlength{\tabcolsep}{1pt}
	\renewcommand{\arraystretch}{0.8}
    \newcommand{\sizea}{0.30\linewidth}
    \footnotesize
	\centering
	\begin{tabular}{cccc}
	  &\emph{t} +  1 & \emph{t} +  3 & \emph{t} +  5 \\
      \rotatebox{90}{\ Sheng \cite{sheng2020high}} &       \includegraphics[width=\sizea]{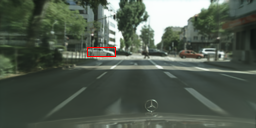} &
      \includegraphics[width=\sizea]{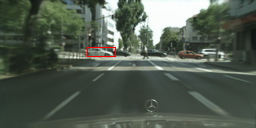} &
      \includegraphics[width=\sizea]{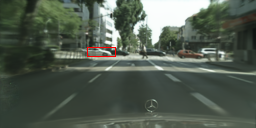}
      \\
	  \rotatebox{90}{\ \ \ \ Sheng*}   &
	  \includegraphics[width=\sizea]{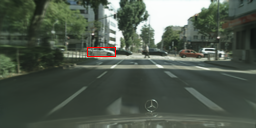} &
      \includegraphics[width=\sizea]{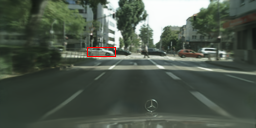} &
      \includegraphics[width=\sizea]{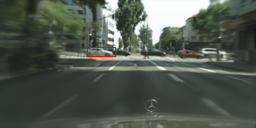}
      \\
	   \rotatebox{90}{\ \ S. Sheng*}&
      \includegraphics[width=\sizea]{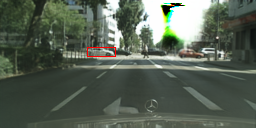} &
      \includegraphics[width=\sizea]{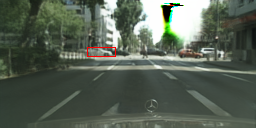} &
      \includegraphics[width=\sizea]{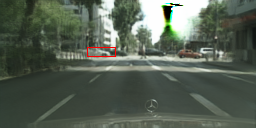}
      \\
	   \rotatebox{90}{\ \ \ \ \ \ Ours} &
      \includegraphics[width=\sizea]{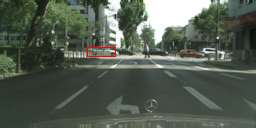} &
      \includegraphics[width=\sizea]{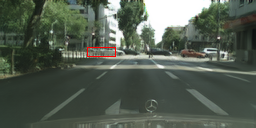} &
      \includegraphics[width=\sizea]{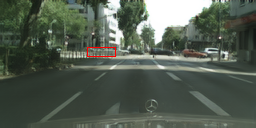}
      \\
	   	   \rotatebox{90}{\ \ \ \ \ GT} &
	  \includegraphics[width=\sizea]{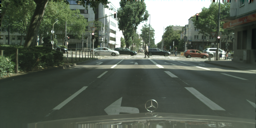} &
      \includegraphics[width=\sizea]{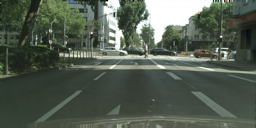} &
      \includegraphics[width=\sizea]{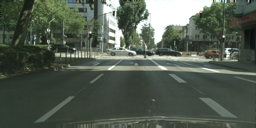}\\
	   
	\end{tabular}
	\caption{Results of predicting the frames \emph{t} + 1,  \emph{t} + 3 , and  \emph{t} + 5 on the Cityscapes dataset~\cite{cordts2016cityscapes} with ground truth reference. On first three columns, we move the pedestrian near the semaphore to left. On the last three columns we move car crossing the street. The position of the moved object at $t=0$ is highlighted in red. Zoom for details.}
	\label{Fig:cityscapes_supplementary2}
\end{figure*}

For the Cityscapes dataset, we train an additional model at higher resolution (i.e. $256\times128$ pixels) without changing any hyper-parameter. The results obtained with this model on two initial frames are shown in \Cref{Fig:cityscapes_supplementary1} and \Cref{Fig:cityscapes_supplementary2}. These results are well in-line the the qualitative results reported in the main paper. We observe that the other methods are not able to move the object (see the red bounding boxes that indicate the initial position of the object). Indeed, the cars are either static, in Sheng \emph{et al.} \cite{sheng2020high} and Sheng*, or blurry, in S. Sheng*.  On the contrary, our approach is able to move the object and generates frames of good quality.

\begin{figure*}[ht]
    \setlength{\tabcolsep}{1pt}
	\renewcommand{\arraystretch}{0.8}
    \newcommand{\sizea}{0.16\linewidth}
    \footnotesize
	\centering
	\begin{tabular}{cccc}
	  &\emph{t} +  1 & \emph{t} +  3 & \emph{t} +  5 \\
      \rotatebox{90}{\ Sheng \cite{sheng2020high}} &
      \includegraphics[width=\sizea]{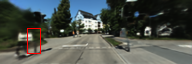} &
      \includegraphics[width=\sizea]{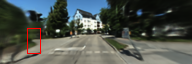} &
      \includegraphics[width=\sizea]{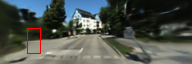}
      \\
	  \rotatebox{90}{\ \ \ \ Sheng*}   &
	  \includegraphics[width=\sizea]{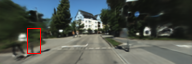} &
      \includegraphics[width=\sizea]{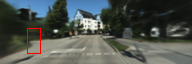} &
      \includegraphics[width=\sizea]{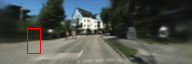}
      \\
	   \rotatebox{90}{\ \ S. Sheng*}&
      \includegraphics[width=\sizea]{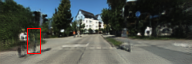} &
      \includegraphics[width=\sizea]{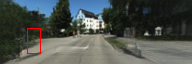} &
      \includegraphics[width=\sizea]{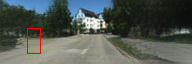}
      \\
	   \rotatebox{90}{\ \ \ \ \ \ Ours} &
      \includegraphics[width=\sizea]{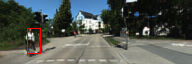} &
      \includegraphics[width=\sizea]{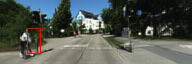} &
      \includegraphics[width=\sizea]{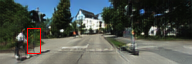}
      \\
	   	   \rotatebox{90}{\ \ \ \ \ GT} &
	   	         \includegraphics[width=\sizea]{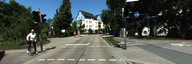} &
      \includegraphics[width=\sizea]{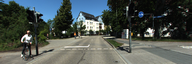} &
      \includegraphics[width=\sizea]{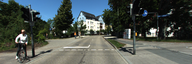}
      \\\\
	   
	\end{tabular}
	\caption{Results of predicting the frames \emph{t} + 1,  \emph{t} + 3 , and  \emph{t} + 5 on the KITTI360 dataset~\cite{Xie2016CVPR}. The position of the moved object at $t=0$ is highlighted in red. Zoom for details.}
	\label{Fig:kitti}
\end{figure*}

\begin{figure*}[ht]
    \setlength{\tabcolsep}{1pt}
	\renewcommand{\arraystretch}{0.8}
    \newcommand{\sizea}{0.16\linewidth}
    \footnotesize
	\centering
	\begin{tabular}{cccc}
	  &\emph{t} +  1 & \emph{t} +  3 & \emph{t} +  5 \\
      \rotatebox{90}{\ Sheng \cite{sheng2020high}} &
      \includegraphics[width=\sizea]{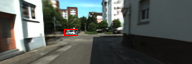} &
      \includegraphics[width=\sizea]{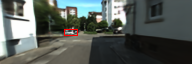} &
      \includegraphics[width=\sizea]{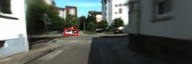}
      \\
	  \rotatebox{90}{\ \ \ \ Sheng*}   &
	  \includegraphics[width=\sizea]{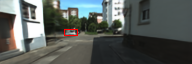} &
      \includegraphics[width=\sizea]{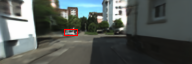} &
      \includegraphics[width=\sizea]{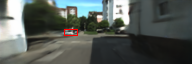}
      \\
	   \rotatebox{90}{\ \ S. Sheng*}&
      \includegraphics[width=\sizea]{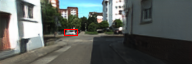} &
      \includegraphics[width=\sizea]{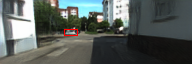} &
      \includegraphics[width=\sizea]{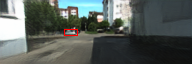}
      \\
	   \rotatebox{90}{\ \ \ \ \ \ Ours} &
      \includegraphics[width=\sizea]{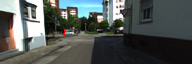} &
      \includegraphics[width=\sizea]{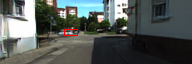} &
      \includegraphics[width=\sizea]{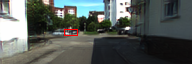}
      \\
	   	   \rotatebox{90}{\ \ \ \ \ GT} &
	   	         \includegraphics[width=\sizea]{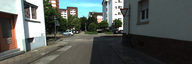} &
      \includegraphics[width=\sizea]{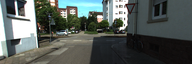} &
      \includegraphics[width=\sizea]{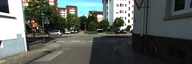}
      \\\\
	   
	\end{tabular}
	\caption{Results of predicting the frames \emph{t} + 1,  \emph{t} + 3 , and  \emph{t} + 5 on the KITTI360 dataset~\cite{Xie2016CVPR}. The position of the moved object at $t=0$ is highlighted in red. Zoom for details.}
	\label{Fig:kitti1}
\end{figure*}

\Cref{Fig:kitti} and \Cref{Fig:kitti1} instead show some additional visual results on KITTI 360.

\begin{figure*}[ht]
    \setlength{\tabcolsep}{1pt}
	\renewcommand{\arraystretch}{0.8}
    \newcommand{\sizea}{0.30\linewidth}
    \footnotesize
	\centering
	\begin{tabular}{cccc}
	  &\emph{t} +  1 & \emph{t} +  3 & \emph{t} +  5 \\
	   
            \rotatebox{90}{\ \ \ \ \ \ (A) Ours} &
      \includegraphics[width=\sizea]{figures/supplementary/cityscapes/ours/lindau_000000_000000_pred_13021_01.png} &
      \includegraphics[width=\sizea]{figures/supplementary/cityscapes/ours/lindau_000000_000000_pred_13021_03.png} &
      \includegraphics[width=\sizea]{figures/supplementary/cityscapes/ours/lindau_000000_000000_pred_13021_05.png}
      \\
      	  \rotatebox{90}{\ \ \ \ (A) w/o GCN.}   &
	  \includegraphics[width=\sizea]{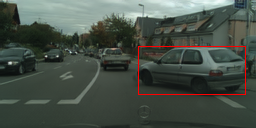} &
      \includegraphics[width=\sizea]{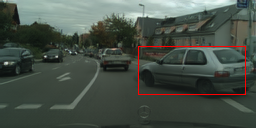} &
      \includegraphics[width=\sizea]{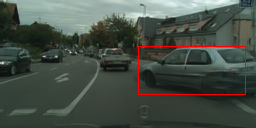}
      \\
	  \rotatebox{90}{\ \ \ \ (A) w/o Obj. Int.}   &
	  \includegraphics[width=\sizea]{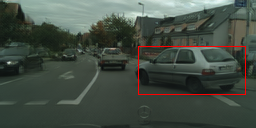} &
      \includegraphics[width=\sizea]{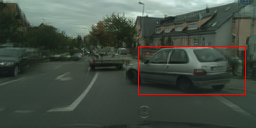} &
      \includegraphics[width=\sizea]{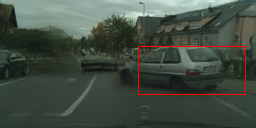}
      \\
      \rotatebox{90}{\ \ \ \ \ \ (A) w/o Sup} &
	  \includegraphics[width=\sizea]{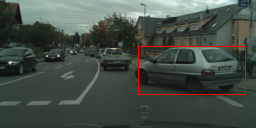} &
      \includegraphics[width=\sizea]{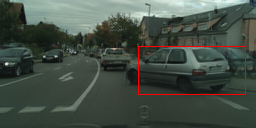} &
      \includegraphics[width=\sizea]{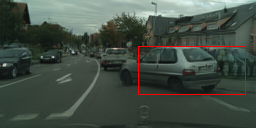}
      \\
	   	   \rotatebox{90}{\ \ \ \ \ \ GT} &
	  \includegraphics[width=\sizea]{figures/supplementary/cityscapes/gt/lindau_000000_000001_leftImg8bit.png} &
      \includegraphics[width=\sizea]{figures/supplementary/cityscapes/gt/lindau_000000_000003_leftImg8bit.png} &
      \includegraphics[width=\sizea]{figures/supplementary/cityscapes/gt/lindau_000000_000005_leftImg8bit.png}\\
	   
	\end{tabular}
	\caption{Results of the ablation test on the Cityscapes dataset~\cite{cordts2016cityscapes}.}
	\label{Fig:cityscapes_supplementary1_ablation}
\end{figure*}

Finally, \Cref{Fig:cityscapes_supplementary1_ablation} shows a qualitative example of the ablation study we performed in the main paper. 
The first row of this Figure shows a version of our network that does not model object interactions. By comparing the first and second rows of \Cref{Fig:cityscapes_supplementary1_ablation}, we clearly see that, while the highlighted object correctly moves, all the other objects have unrealistic motions. This confirms the quantitative results about the importance of modelling the object interactions to have temporal consistent movements of different objects in the scene.

\begin{figure*}[ht]
    \setlength{\tabcolsep}{1pt}
	\renewcommand{\arraystretch}{0.8}
    \newcommand{\sizea}{0.30\linewidth}
    \footnotesize
	\centering
	\begin{tabular}{cccc}
	  &\emph{t} +  1 & \emph{t} +  3 & \emph{t} +  5 \\
      \rotatebox{90}{\ Sheng \cite{sheng2020high}} &       \includegraphics[width=\sizea]{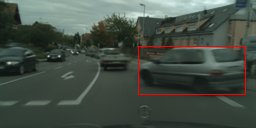} &
      \includegraphics[width=\sizea]{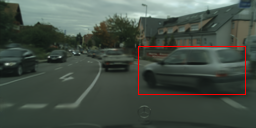} &
      \includegraphics[width=\sizea]{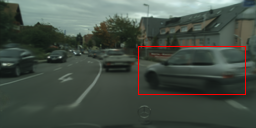}
      \\
	  \rotatebox{90}{\ \ \ \ Sheng*}   &
	  \includegraphics[width=\sizea]{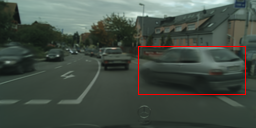} &
      \includegraphics[width=\sizea]{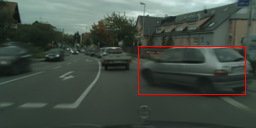} &
      \includegraphics[width=\sizea]{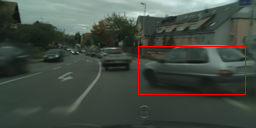}
      \\
	   \rotatebox{90}{\ \ S. Sheng*}&
      \includegraphics[width=\sizea]{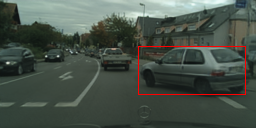} &
      \includegraphics[width=\sizea]{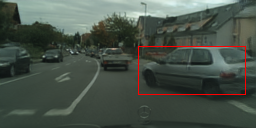} &
      \includegraphics[width=\sizea]{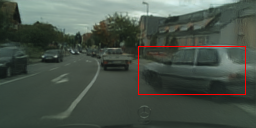}
      \\
	   \rotatebox{90}{\ \ \ \ \ \ Ours} &
      \includegraphics[width=\sizea]{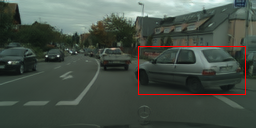} &
      \includegraphics[width=\sizea]{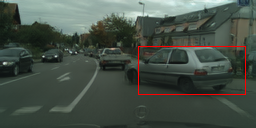} &
      \includegraphics[width=\sizea]{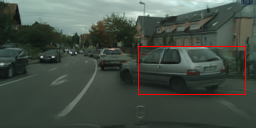}
      \\
	   	   \rotatebox{90}{\ \ \ \ \ \ GT} &
	  \includegraphics[width=\sizea]{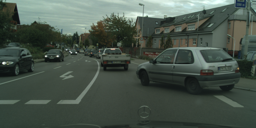} &
      \includegraphics[width=\sizea]{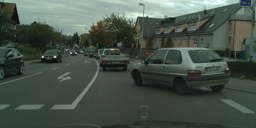} &
      \includegraphics[width=\sizea]{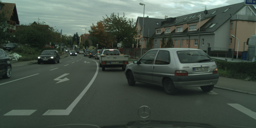}\\
	   
	\end{tabular}
	\caption{Results of predicting the frames \emph{t} + 1,  \emph{t} + 3 , and  \emph{t} + 5 on the Cityscapes dataset~\cite{cordts2016cityscapes} with ground truth reference. On first three columns, we move the pedestrian near the semaphore to left. On the last three columns we move car crossing the street. The position of the moved object at $t=0$ is highlighted in red. Zoom for details.}
	\label{Fig:cityscapes_supplementary3}
\end{figure*}

\end{document}